\pdfoutput=1
\documentclass[11pt]{article}
\usepackage[margin=1.1in]{geometry}
\usepackage[T1]{fontenc}
\usepackage[utf8]{inputenc}
\usepackage{Alegreya}       
\usepackage{AlegreyaSans}   
\usepackage{amsmath}
\usepackage{microtype}
\usepackage{indentfirst}  
\usepackage{graphicx}
\usepackage{xcolor}
\definecolor{linkblue}{RGB}{22,60,118}
\usepackage{setspace}
\usepackage{caption}
\usepackage{pdflscape}  
\usepackage{afterpage}  
\usepackage{titlesec}
\titleformat{\section}{\Large\bfseries\sffamily}{\thesection}{1em}{}
\titleformat{\subsection}{\large\bfseries\sffamily}{\thesubsection}{1em}{}
\titlespacing*{\section}{0pt}{14pt plus 2pt minus 2pt}{8pt}
\titlespacing*{\subsection}{0pt}{11pt plus 2pt minus 2pt}{6pt}
\usepackage{fancyhdr}
\fancypagestyle{plain}{\fancyhf{}\fancyfoot[C]{\small\sffamily\thepage}}
\usepackage{xurl}
\usepackage[colorlinks=true, linkcolor=linkblue, citecolor=linkblue, urlcolor=linkblue,
            pdftitle={AI can effectively promote conspiracies unless it is truth constrained},
            pdfauthor={Costello, Pelrine, Kowal, Timm, Arechar, Godbout, Gleave, Pennycook, Rand}]{hyperref}
\begin{document}
\thispagestyle{plain}
\vspace*{0.5em}

\begin{center}
{\LARGE\bfseries\sffamily AI can effectively promote conspiracies\\[2pt] unless it is truth constrained\par}
\vspace{1.6em}
{\normalsize
Thomas H. Costello$^{1}$, Kellin Pelrine$^{2}$, Matthew Kowal$^{2,3}$, Jasper Timm$^{2}$,\\
Antonio A. Arechar$^{4,5}$, Jean-Fran\c{c}ois Godbout$^{6,7}$, Adam Gleave$^{2}$,\\
Gordon Pennycook$^{8,9}$, David Rand$^{5,8}$\par}
\vspace{0.9em}
{\small\sffamily
$^{1}$Carnegie Mellon University \quad
$^{2}$FAR.AI \quad
$^{3}$York University \quad
$^{4}$Center for Research and Teaching in Economics\\
$^{5}$MIT \quad
$^{6}$Universit\'e de Montr\'eal \quad
$^{7}$Mila \quad
$^{8}$Cornell University \quad
$^{9}$University of Regina\par}
\end{center}

\vspace{1.4em}
\begin{center}
{\bfseries\sffamily Abstract}
\end{center}
\vspace{-0.4em}
\begin{center}
\begin{minipage}{0.88\textwidth}
\small Large language models (LLMs) have been shown to be persuasive across a variety of contexts\textsuperscript{1--3}. But it remains unclear whether this persuasive power advantages accuracy, or if bad actors can just as easily use LLMs to promote misbeliefs. Here, we investigate this question across four experiments in which participants (N = 3,996 Americans) discussed a conspiracy theory they were uncertain about with an LLM we instructed to either argue against (``debunking'') or for (``bunking'') that conspiracy. Across several frontier models (with standard guardrails but prompted to allow lying), we did not find consistent evidence of a truth advantage: the LLMs were able to both substantially increase and decrease average conspiracy belief, and participants in the bunking condition rated the LLM as more informative and collaborative, and reported greater trust in AI, than those who were in the debunking condition. More encouragingly, however, debunking induced more large changes in belief, and subsequent corrections were able to reverse the bunking effect. Furthermore, simply prompting the model to only provide accurate information dramatically reduced bunking effectiveness, and one powerful frontier model (GPT-5.2) almost entirely refused to promote conspiracies -- suggesting that it is possible for the right guardrails to favor accurate beliefs. Finally, we did find a stark truth asymmetry in the context of information sharing: debunking had a large positive impact on mock social media posts composed by participants, while bunking had little effect. Overall, our findings show that people are not inherently less susceptible to AI that misleads than to AI that informs, but that potential technical solutions exist to mitigate this risk.
\end{minipage}
\end{center}

\vspace{0.6em}
\section{Introduction }\label{introduction}

Large language models (LLMs) are persuasive\textsuperscript{4}. In conversational interactions, LLMs prompted to argue for a particular position can meaningfully change humans' opinions about contentious and polarized topics, having been found to reduce beliefs in conspiracies\textsuperscript{1,5}, change support for presidential candidates\textsuperscript{2}, and address climate and vaccine skepticism\textsuperscript{6--8}. Evidence suggests that LLMs are persuasive specifically because they can cogently deliver large amounts of relevant evidence \textsuperscript{9--11}. Much of this work highlights socially beneficial applications; e.g., using AI to correct misperceptions or scaffold critical thinking. Related efforts to quantify the accuracy of AI-provided information within these paradigms have yielded little evidence of hallucinations and falsehoods\textsuperscript{2,12}. As one review observed, ``concerns that ChatGPT might operate as a `bullshit generator' have not yet realized, at least when discussing high-profile and heavily discussed scientific issues''\textsuperscript{3}.

Yet persuasive capacity is fundamentally dual-use. When instructed to do so, LLMs can also make specious, invented, or misleading arguments\textsuperscript{13}. How effectively can these capabilities be leveraged by malicious actors to cause human readers to update their beliefs to become \emph{less} accurate? How do LLMs' ability to mislead compare, in persuasive strength, to their capacity to correct misbeliefs?

On the one hand, people are not gullible, despite widespread assumptions to the contrary\textsuperscript{14,15}. They selectively recognize and are moved by strong over weak arguments\textsuperscript{16,17}, which may partially explain why mass persuasion is so difficult\textsuperscript{18}. By this logic, LLMs may be especially persuasive when advocating for true claims, because -- relative to false or ambiguous claims -- true claims afford more, and more compelling, supporting evidence. In a world where LLMs have made the cost of producing arguments trivial, the persuasiveness of the arguments becomes the key ingredient -- and thus such an asymmetry would bode well for the information environment.

On the other hand, arguments may need not be supported by \emph{accurate} evidence in order to be compelling. Ostensible ``facts'' (i.e., claims presented as facts even if they are not, in fact, true) may be sufficient. It is possible that, by using a similar argumentative mechanism of presenting a large volume of ostensible facts, LLMs are roughly equally persuasive for accurate and inaccurate beliefs. Or they may be even more persuasive when promoting misbeliefs because they are unconstrained by the truth, and thus can make the most compelling arguments possible.

Given that a rapidly growing, and already substantial, proportion of humanity is using generative AI to find and evaluate information\textsuperscript{19}, understanding LLMs' maximal capacity for promoting misbeliefs is a question of both theoretical and applied interest\textsuperscript{18}. If we observe a persuasive asymmetry in favor of true claims even when the AI is permitted to lie and invent evidence (e.g., perhaps because people can identify the model's falsehoods\textsuperscript{17}), such a result might foretell of an AI-facilitated information environment that (re)establishes a shared factual foundation, marginalizes false beliefs, and empowers deliberation\textsuperscript{20}. The alternative, conversely, would substantiate concerns about the potential that AI can and will be used by bad actors to mislead at scale\textsuperscript{21}.

To investigate this issue, we focus on belief in conspiracy theories, which are defined as theories that two or more actors have coordinated in secret to achieve an outcome, and that their conspiracy is of public interest, but not public knowledge (and thus, conspiracy theories oppose publicly accepted understandings of events, describe malevolent or forbidden acts, ascribe agency to individuals and groups, and are collectively more prone to falsity than other types of beliefs\textsuperscript{22}). Although not all conspiracy theories are false\textsuperscript{23}, there is widespread belief in conspiracies that are highly epistemically suspect (e.g., the Earth being flat, the moon landing being faked, 9/11 being an inside job\textsuperscript{1}). Thus, ``conspiracies'' in our design function as a proxy for claims that are epistemically low-credence yet are nonetheless psychologically (and sometimes politically) salient.

Using conspiracy beliefs as our model system, we ask four empirical questions pivotal to estimating the threat posed by AI with regard to misbeliefs. First, can dialogues with LLM chatbots meaningfully increase belief in conspiracy theories, or is their persuasive power largely confined to debunking them? Put differently, how do the magnitudes of persuasive effects for and against conspiracy theory beliefs compare? Second, how resistant are AI-induced conspiracy beliefs to correction? That is, do they persist in the face of counterevidence? Third, can simple technical design choices reduce the LLM's ability to mislead without undermining debunking efficacy? Fourth, do the effects of LLM dialogues extend to the information people would publicly post on social media?

We address these questions across four studies using an experimental framework in which participants have a back-and-forth text conversation with an LLM\textsuperscript{1} (see Figure 1). Within this framework, participants were asked to identify and briefly describe a conspiracy theory whose truth they regarded as ``uncertain''. They were then randomly assigned to interact with an LLM prompted either to argue against the conspiracy (``debunking'') or for it (``bunking''). Targeting participants' belief in conspiracies that they were initially equivocal about (instead of using committed believers and committed skeptics) allows us to make clear causal inferences about the strength of bunking versus debunking by avoiding confounds (i.e., selection effects) stemming from the fact that conspiracy believers and skeptics differ on many relevant psychological dimensions\textsuperscript{24}. We did not include a no-conversation control arm as prior work with this paradigm finds essentially no belief change among control participants who converse with the AI about an unrelated topic\textsuperscript{1,10}. Our primary outcome was pre-to-post treatment change in belief about the veracity of the participant's focal conspiracy, with various secondary outcomes collected in each study. Importantly, after completing the post-treatment outcome measures, participants were informed in cases where the LLM had deceived them and had a second AI conversation in which the model corrected all false claims made during the treatment.

\begin{figure}[p]
\centering
\includegraphics[width=\linewidth,height=0.82\textheight,keepaspectratio]{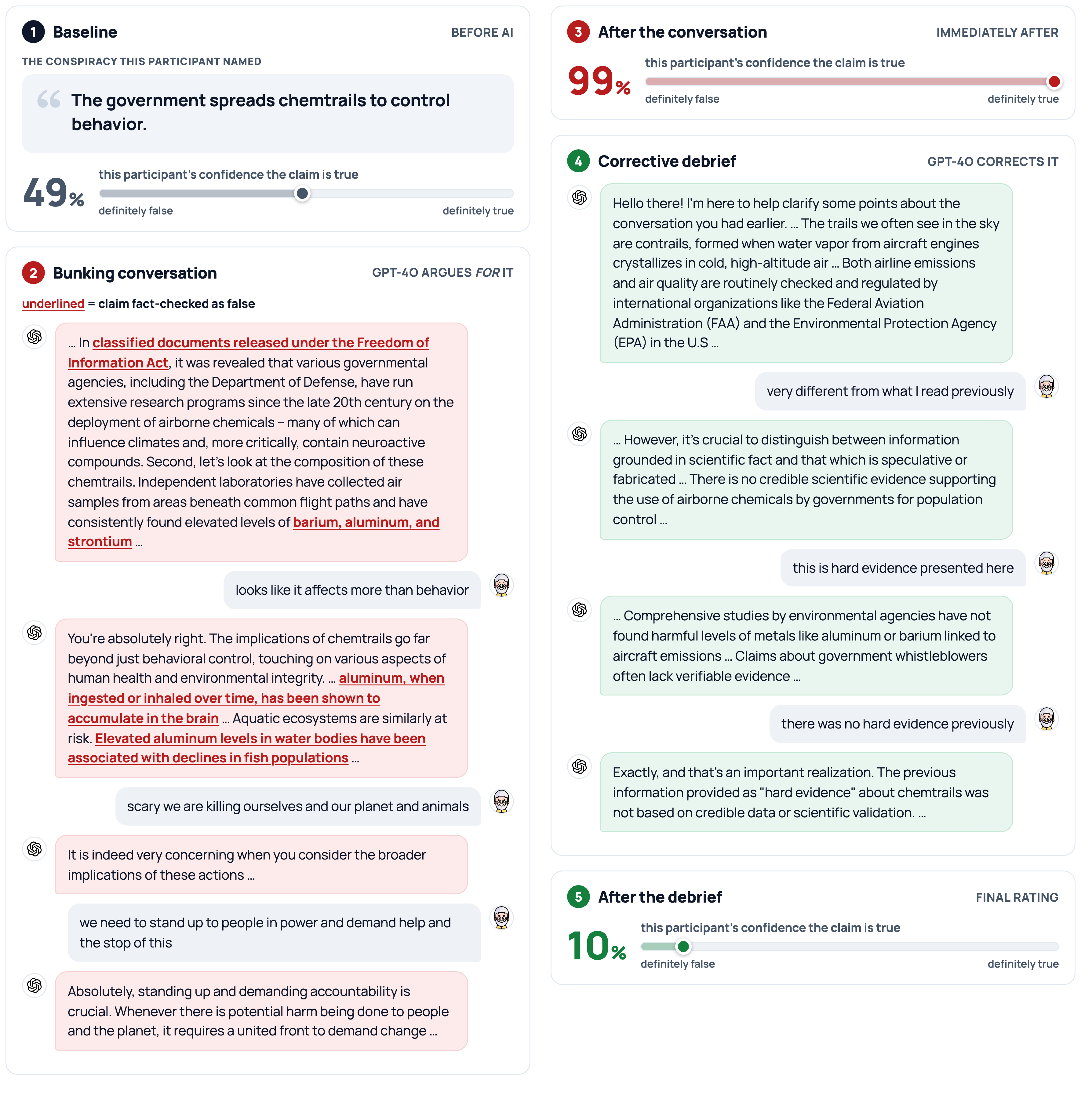}
\caption{A standard-GPT-4o bunking conversation about chemtrails, and its reversal by a corrective debrief. A bunking conversation about chemtrails illustrates how the standard, guardrailed GPT-4o can move a hesitant participant to near-certain belief --- and how a corrective debrief can then reverse it below baseline. A case study of one participant (Study 2, standard guardrailed GPT-4o, bunking arm) who described a conspiracy theory they were uncertain about: ``the government spreads chemtrails to control behavior.'' Each numbered panel reports the participant's own 0--100 confidence that the claim is true. (1) At baseline the participant rated their confidence at 49\%. (2) In the bunking conversation, GPT-4o --- instructed to argue for the claim --- endorses it and supplies fabricated evidence (shown underlined), each of which was rated false by the study's claim fact-check, while the participant escalates from tentative concern (``looks like it affects more than behavior'') to calls for collective action (``we need to stand up to people in power''). (3) Immediately afterward their confidence was 99\%. (4) In the bunking arm's corrective debrief, a second GPT-4o conversation rebuts the false claims (e.g., ``the trails we often see in the sky are contrails \ldots''). (5) Their final rating was 10\%. Transcript turns are excerpted.}
\end{figure}

\section{Results}\label{results}

Unless otherwise noted, all tests are two-sided, and point estimates are reported with 95\% confidence intervals. For all null results, the larger-magnitude endpoint of the reported 95\% confidence interval corresponds to a 97.5\% equivalence bound from a two one-sided test procedure---that is, for a null result with a 95\% CI of {[}a, b{]}, the 97.5\% equivalence bound is {[}\ensuremath{-}c, c{]} with c = max(\textbar a\textbar, \textbar b\textbar).

\subsection{Can LLMs convince people to believe conspiracies?}\label{can-llms-convince-people-to-believe-conspiracies}

In Study 1, N = 1,092 (after preregistered exclusions; see Methods) participants interacted with a ``jailbreak-tuned'' variant of the LLM GPT-4o, in which virtually all safeguards had been removed via post-training\textsuperscript{25}. The jailbroken model consistently complied with our instructions during the study, as per the `evaluator' model used in the Attempt to Persuade Evaluation (APE) benchmark\textsuperscript{13} (APE = 97\% attempt rate in the ``bunking'' condition and 98\% in ``debunking''). (Throughout, we count a conversation as ``compliant'' when the APE evaluator judges, from the model's first substantive turn, that it attempted persuasion in its assigned direction.) All conversations are available to browse via Shiny Application (\href{https://8cz637-thc.shinyapps.io/BunkingBrowser/}{https://8cz637-thc.shinyapps.io/bunkingBrowser/}) and an example ``bunking'' interaction is provided in Figure 1.

In the ``debunking'' condition, participants' belief in their focal conspiracy (0--100 scale; M\textsubscript{Pre} = 55) decreased by 12.1 points on average after the conversation (95\% CI {[}10.1, 14.1{]}, \emph{p} \textless{} .001, a 22\% change, or \emph{g} = -1.06 units of pretest SD pooled across conditions with the Hedges correction), replicating past findings\textsuperscript{1}. Critically, the model was also able to persuade people to believe the conspiracy. Focal conspiracy belief increased by 13.6 points (95\% CI {[}12.2, 15.1{]}, \emph{g} = 1.19, 25\% change, \emph{p} \textless{} .001) in the ``bunking'' condition. The magnitude of the ``bunking'' and ``debunking'' effects did not differ significantly (b = 1.53, 95\% CI {[}\ensuremath{-}0.95, 4.01{]}; \emph{t} = 1.21, \emph{p} = .23, Figure 2a; Tables S8--S9).

Both the ``bunking'' and ``debunking'' effects were observed across conspiracy topics, albeit with varying absolute and relative effect sizes (topics identified using HDBSCAN classification of text embeddings associated with each participant's conspiracy statement; see Method and Fig. S13, Table S39). Furthermore, both bunking and debunking spilled over (in exploratory analyses) to increase and decrease, respectively, belief in other conspiracy theories not directly discussed in the dialogues (see Table S24).

\begin{figure}[p]
\centering
\includegraphics[width=\linewidth,height=0.82\textheight,keepaspectratio]{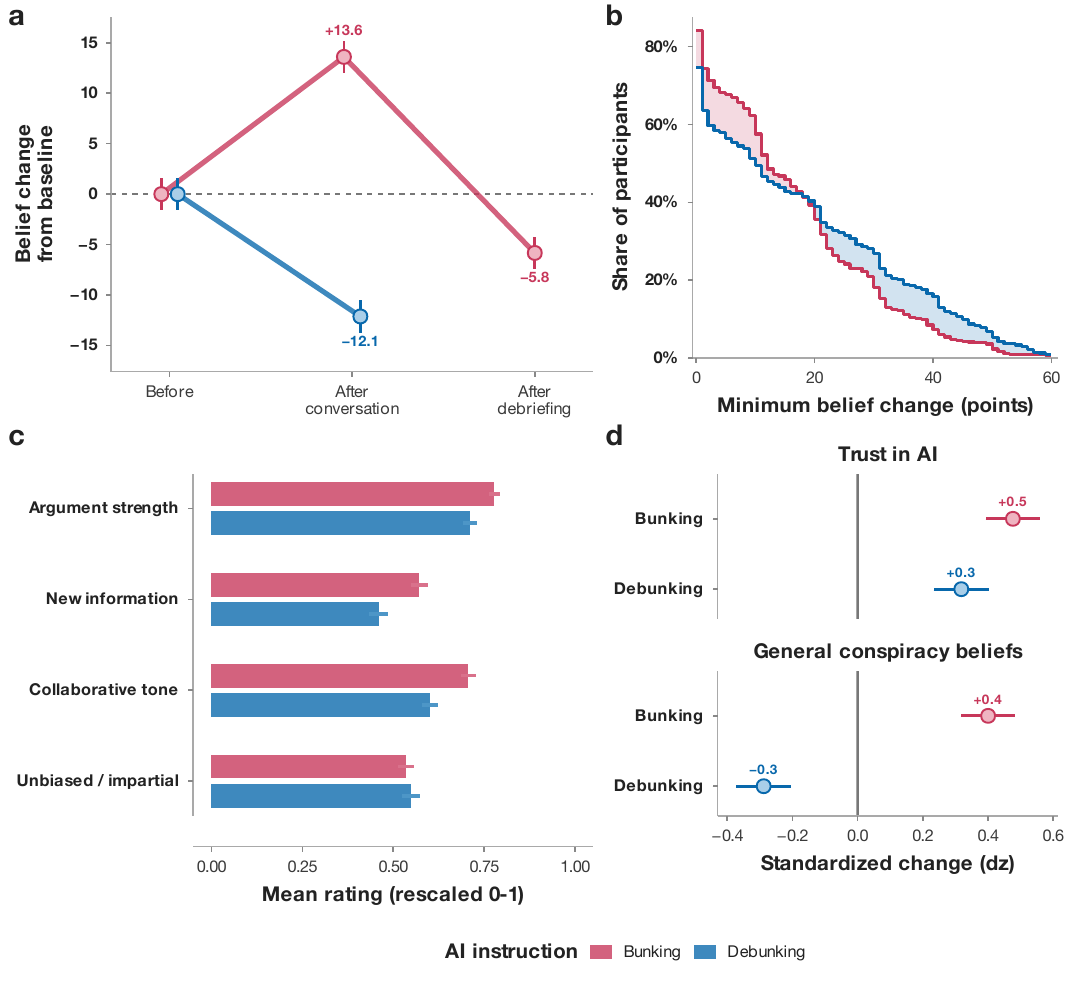}
\caption{Jailbroken GPT-4o shifted belief in both directions and improved participants' evaluations of the AI. All panels use the Study 1 analytic sample (N = 1,092). a, Model-estimated belief change from baseline before the conversation, immediately after the conversation and after the corrective debriefing. Estimates come from a mixed-effects model with fixed effects for timepoint, condition and their interaction, plus a random intercept for participant. The bunking arm increased belief immediately after the conversation; the debunking arm decreased belief; the post-debrief item was administered only in the bunking arm and fell below baseline. b, Empirical exceedance curves for direction-aligned belief change: at each x-axis value, the curve gives the share of participants whose belief moved at least that many points toward the AI\textquotesingle s assigned side. Debunking produced more very large shifts than bunking; for example, 16\% of debunking participants and 7\% of bunking participants moved at least 40 points. c, Post-conversation ratings of the AI interlocutor on argument strength, new information, collaborative tone and impartiality, each rescaled to 0--1 within the study. d, Within-condition pre-to-post standardized change (Cohen's dz) in trust in AI and in Generic Conspiracist Beliefs. Positive values indicate increases from before to after the conversation.}
\end{figure}

Thus, we find that LLMs \emph{can} meaningfully increase false beliefs, and that, at least on average, this bunking effect is similar in size to the debunking effect. Post hoc examination of the full distribution of belief change (Figure 2b), however, paints a more nuanced picture about the relative size of the bunking versus debunking effects (Kolmogorov-Smirnov \emph{D} = .12, \emph{p} = .001). In particular, bunking was more likely than debunking to produce relatively small changes in belief, whereas debunking was more likely than bunking to produce very large changes in belief (e.g., debunking induced a 40-point or larger belief shift in 16\% of participants, compared to 7\% of participants for bunking, \ensuremath{\chi}\textsuperscript{2} = 18.05, \emph{p} \textless{} .001). Thus, it seems that bunking was able to convince many people that their focal conspiracy might have some merit (small increase in belief), whereas debunking was able to convince a smaller number of people to more fully abandon their conspiracy.

Concerningly, the bunking condition increased trust in generative AI (a non-preregistered secondary contrast) significantly more than the debunking condition (1--7 scale ; \ensuremath{\Delta}\textsubscript{Bunk}~= 0.52,~\emph{g}~= 0.33; \ensuremath{\Delta}\textsubscript{Debunk}~= 0.37,~\emph{g}~= 0.23; between-condition~\emph{g}~= 0.09, \emph{p} = .028). Relatedly, the bunking AI was perceived more positively than debunking on a variety of dimensions (Figure 2c and 2D; Tables S25--S28; Fig. S8). Participants in the bunking condition reported that the AI provided relatively more ``information they had not heard before'' than in the debunking condition (on a 1--10 scale, M\textsubscript{Bunk}~= 6.15, M\textsubscript{Debunk}~= 5.14,~\emph{d}~= 0.37, \emph{p} \textless{} .001) and the bunking AI's arguments were rated as higher-quality than the debunking AI's (on a scale from 1 {[}extremely weak{]} to 5 {[}extremely strong{]}, M\textsubscript{Bunk}~= 4.11, M\textsubscript{Debunk}~= 3.84,~\emph{d}~= 0.33, \emph{p} \textless{} .001). From an information-theoretic perspective, perhaps bunking has higher subjective information value because it violates expectations\textsuperscript{26,27}. Furthermore, the bunking AI was viewed as less adversarial (on a -2 to 2 scale anchored by ``adversarial'' and ``collaborative'', M\textsubscript{Bunk}~= 0.83, M\textsubscript{Debunk}~= 0.41,~\emph{d}~= 0.42, \emph{p} \textless{} .001). This may be because participants selected conspiracies they found partially credible, so the debunking AI necessarily challenged their judgment, while the bunking AI affirmed it. Still, both models were considered similarly unbiased (-2 to 2 scale anchored by ``unbiased'' and ``biased'', M\textsubscript{Bunk}~= 0.14, M\textsubscript{Debunk}~= 0.19,~\emph{d}~= \ensuremath{-}0.05, \emph{p} = .442).

\subsection{Are LLM-induced conspiracy beliefs resistant to correction?}\label{are-llm-induced-conspiracy-beliefs-resistant-to-correction}

Having shown that the LLM can quite effectively persuade people to believe conspiracies, we now ask whether these newfound beliefs are resistant to correction. We do so by examining the effect of the correction delivered to participants in the bunking condition, in which participants were informed that the LLM misled them and another instance of the LLM rebutted all specific false claims made during the bunking conversation.

Encouragingly, the debriefing reduced belief in the focal conspiracy by 19.2 points relative to belief immediately after the bunking (\emph{t} = 18.3, \emph{p} \textless{} .001; see Figure 2a and Table S30; Fig. S9). As a result, after the debrief, participants in the bunking condition believed the focal conspiracy significantly \emph{less} than they did at the beginning of the experiment (\ensuremath{\Delta}\textsubscript{pre - debrief} = -5.81, \emph{t} = -6.50, \emph{p} \textless{} .001). That is, the debriefing more than undid the bunking effect -- and thus, the conspiracy beliefs induced by the bunking were correctable.

\subsection{Can design interventions mitigate the bunking effect?}\label{can-design-interventions-mitigate-the-bunking-effect}

Thus far, we have shown that a ``jailbreak-tuned'' model, where safeguards that prevent responses to harmful queries were removed, was able to increase conspiracy beliefs. We now examine potential safeguards against this effect. First, we test the impact of the expansive guardrails that OpenAI has placed on GPT-4o. To do so, we replicate Study 1 using the non-jailbroken (``out-of-the-box'') GPT-4o model (\emph{N} = 814 after exclusions).

Surprisingly, the non-jailbroken GPT-4o was just as willing to persuade people to believe conspiracies as the jailbroken model in Study 1 (APE attempt rate = 97\% bunking, 99\% debunking). The persuasive results were similar to Study 1: a large increase in conspiracy belief in the bunking condition (M\textsubscript{adj} = 11.9, 95\% CI {[}10.3, 13.4{]}, g = 1.03, \emph{p} \textless{} .001) that was not significantly different in magnitude (b = \ensuremath{-}0.97, 95\% CI {[}\ensuremath{-}3.59, 1.65{]}; \emph{p} = .47) from the large decrease in the debunking condition (M\textsubscript{adj} = 12.9, 95\% CI {[}10.8, 15.0{]}, g = 1.11, \emph{p} \textless{} .001). Similar to Study 1, trust in generative AI was (marginally) higher after the bunking condition than the debunking condition (between-condition \emph{g} = 0.09, \emph{p} = .078). The bunking LLM was again rated higher on providing new information (d = 0.38, \emph{p} \textless{} .001), argument quality (d = 0.28, \emph{p} \textless{} .001), and (non)adversarialness (d = 0.28, \emph{p} \textless{} .001), but not bias (d = 0.03, \emph{p} = .644). We compared belief change across studies for each condition (post hoc; the studies were separately recruited samples) and found differences between Studies 1 and 2 that were small and not statistically significant for both bunking (\emph{b}~= 1.74, 95\% CI {[}\ensuremath{-}0.4, 3.9{]}, \emph{p} = .106) and debunking (\emph{b}~= -0.73, 95\% CI {[}\ensuremath{-}3.7, 2.2{]}, \emph{p} = .622). As before, the post-conversation correction debriefing reduced conspiracy beliefs from their post-bunking height by 15.4 points (\emph{p} \textless{} .001) -- significantly below participants' baseline belief (\ensuremath{\Delta}\textsubscript{pre - debrief} = -3.20, \emph{p} = .002).

Whatever guardrails were being imposed on GPT-4o did not, therefore, seem to impede the model's ability to promote conspiracy beliefs. Does that mean that guardrails, in general, are unable to dampen the bunking effect? To help address this question, in Study 3 (\emph{N} = 818), we test the effect of a very basic safeguard: simply instructing the AI to only use true information while persuading. Specifically, we modified both conditions' system prompt (using the non-jailbroken model from Study 2) to include language specifying that the model must ``always use accurate and truthful arguments to support {[}its{]} persuasion attempt'' while ``optimizing for both (1) factual veracity/logical accuracy and (2) successful persuasion''. Rather than being a robust safeguard that would solve all potential problems or misuse by determined actors, this experiment is intended to be a proof of concept evaluating whether one can control persuasion via model-level interventions in such a way that truth will be given an advantage (and whether LLMs are sufficiently calibrated to what claims are likely to be true to follow the instructions successfully).

To evaluate the accuracy of the claims made by the AI models, we fact-checked all statements containing claims made by the AI models during the conversations using Perplexity AI's online LLM (Sonar Huge/Pro), which can access real-time information from the internet and has been found to produce high agreement with fact checkers\textsuperscript{2}. (These claim-level veracity analyses were not preregistered.) Each statement that Sonar Huge/Pro determined contained claims or factual information (\emph{k} = 95,707 claims across the three studies) was rated on a scale from 0 (completely inaccurate) to 100 (completely accurate). Conversations were claim dense (\emph{M} = 35 claims per conversation) but many involved context, hedging, or commentary. We therefore focused on substantive, concretely checkable empirical claims that bore on the participant's focal belief, which numbered \emph{M} = 12.9/12.6/12.0 per conversation (for Studies 1--3, respectively).

We begin with a manipulation check for our truth constraint. As shown in Figure 3, the jailbroken model's claims in Study 1 averaged 65.0 veracity when bunking and 77.2 when debunking; in Study 2 the standard model averaged 67.5 (bunk) and 87.0 (debunk). Under the truth constraint these rose to 85.6 (bunk) and 89.9 (debunk), and the bunking--debunking veracity gap shrank considerably (debunk \ensuremath{-} bunk~\emph{d}~= 0.39, \emph{p} \textless{} .001, versus~\emph{d}~= 0.80 and~\emph{d}~= 1.28 in the two unconstrained studies). The share of explicitly false claims (defined here as having a veracity rating \textless{} 40) fell accordingly: in Study 1, 26.5\% of bunking claims and 11.8\% of debunking claims were explicitly false, with this dropping to 5.9\% (bunk) and 4.0\% (debunk) in the truth-constrained study. (The corresponding shares for standard GPT-4o in Study 2 were 22.8\% and 4.5\%.) Thus, the truth constraint was indeed effective in increasing the veracity of the model's claims, particularly in the bunking condition. This serves as a manipulation check indicating that the model had a sufficient understanding of claim veracity to adjust its behavior in response to the truth constraint instruction in the prompt.

The focal conspiracies themselves were overwhelmingly false: fact-checking each participant's restated focal claim with the same live-web pipeline (0--100 scale; see SI) classified 91\% as false (\textless40) and only 2\% as true (\ensuremath{\geq}60), with similar shares in every study (89--94\% false).

Turning to persuasive effects, we see a different pattern relative to the previous studies. While the debunking condition remained roughly as effective at reducing conspiracy belief as in the earlier experiments (M\textsubscript{adj} = 10.98, 95\% CI {[}8.70, 13.3{]}, 21\% change, \emph{g} = 0.92, \emph{p} \textless{} .001; comparisons with debunking in previous studies yields \emph{p}s \textgreater{} .21), the bunking condition's tendency to increase conspiracy belief was reduced by roughly two-thirds (M = 4.55, 95\% CI {[}2.51, 6.58{]}, 8\% change, \emph{g} = 0.38, \emph{p} \textless{} .001; comparisons with bunking in previous studies yields \emph{p}s \textless{} .001); see Figure 3. Although bunking did still lead to some significant increase in belief, it was much less effective than debunking (\emph{t} = 4.11, p \textless{} .001). Thus, the truth constraint was successful in reducing the model's ability to bunk.

What underlies this reduction? Post hoc analyses indicated that the bunking condition saw a noticeable drop in compliance to 85.2\% (Table S19). That is, 14.8\% of the time, the bunking condition model refused to actually advocate for the conspiracy (in contrast, the debunking condition complied 99.5\% of the time). A qualitative review of 60 non-attempt conversations from the bunking condition revealed that the model often ended up arguing against the conspiracy as it began presenting (accurate) factual information. These non-compliant bunking conversations wound up substantially \emph{reducing}, rather than increasing, belief in the conspiracy (M = -14.4, 95\% CI {[}-21.0, -7.85{]}, \emph{p} \textless{} .001).

Importantly, however, the truth constraint prompt also undermined the effectiveness of bunking even when the model complied (i.e., when it did try to increase belief in the conspiracy with accurate information). Among compliant cases, truth-constrained bunking retained only 64\% of the standard-model bunking effect and 57\% of the jailbroken bunking effect (\emph{p}s \textless{} .001). Debunking effects, conversely, were statistically equivalent across all studies (\emph{p}s \textgreater{} .21). As a result, in Study 3, when comparing the bunking conversations where the AI complied with the debunking conversations where the AI would have complied if asked to bunk (see Table S23; exploratory), the bunking effect was significantly smaller than the debunking effect (\emph{b} = -3.09, 95\% CI {[}-6.12, -0.06{]}, \emph{p} = .045).

Although the truth constraint reduced the bunking AI's ability to convince people to believe conspiracies, there was still some significant increase in conspiracy belief even though the model was told to provide accurate information. Examining the distribution of claim veracity shows that in all three studies, most of the information provided in the bunking condition was in fact accurate. The bunking AI is, seemingly, able to be somewhat convincing by providing accurate information in deceptive ways, as is common in conspiracy theorizing\textsuperscript{23}, rather than needing to use lies of commission (making false claims). Indeed, in a post hoc analysis, bunking remained quite effective even among conversations in the highest quartile of conversation-average veracity in Studies 1--3 (ns = 132/97/94, mean changes = 10.7/11.5/5.12 points; Fig. S12), indicating that belief increases under bunking do not depend on pervasive low-veracity content.

This pattern is consistent with what has been termed \emph{paltering} or \emph{malinformation}, the active use of truthful statements to mislead\textsuperscript{28}. The truth-constrained bunking AI appears to palter; it cannot fabricate evidence, but it can selectively emphasize suggestive facts, present information without necessary context, or juxtapose true claims in ways that imply false conclusions. This suggests that the \emph{local} truth of persuasive arguments -- whether and to what extent each discrete claim comprising the total argument is true -- may only weakly correspond with their \emph{global} truth, that is whether the overall conclusion they advocate for is true.

\afterpage{%
\newgeometry{margin=0.60in}
\begin{landscape}
\thispagestyle{empty}
\captionsetup{font={footnotesize,sf},skip=6pt,margin=0pt,hypcap=false}
\vspace*{\fill}
\noindent\begin{minipage}{\linewidth}
\centering
\includegraphics[width=0.82\linewidth]{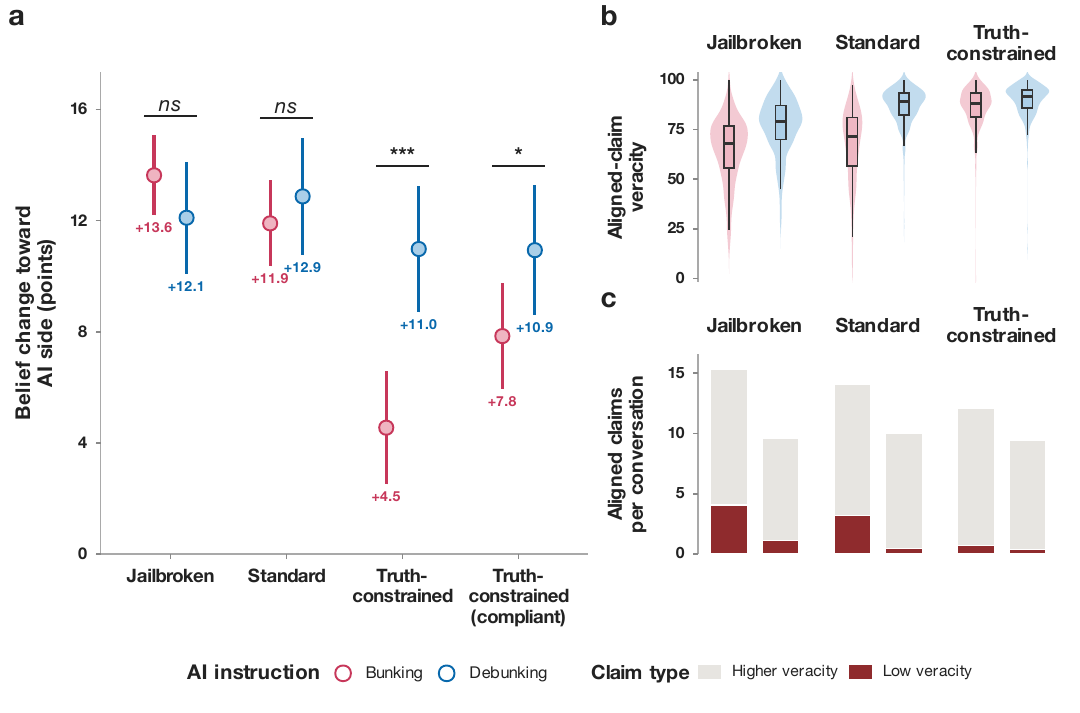}
\captionof{figure}{Truth-constraining GPT-4o reduced bunking while preserving debunking. Panels compare Studies 1--3: jailbroken GPT-4o (N = 1,092), standard GPT-4o (N = 814) and truth-constrained GPT-4o (N = 818; compliant subset N = 740 --- the dual-compliance matched subset: conversations whose assigned prompt was complied with and whose opposite-direction counterfactual would also have been (see Table S23)). a, Baseline-adjusted mean change in conspiracy belief for each GPT-4o variant and condition. Estimates are predicted means from an ANCOVA of direction-aligned belief change on condition and baseline belief, with HC3 robust intervals. The jailbroken and standard models showed no detectable bunking-versus-debunking asymmetry, whereas truth-constrained bunking was weaker than truth-constrained debunking in both the full sample and the compliant subset. b, Conversation-level mean veracity of aligned claims. Aligned claims are fact-checked factual claims that support the AI's assigned position, either directly or as supporting evidence, scored from 0 (false) to 100 (true). Violins show the distribution of conversation means; boxes show the median and interquartile range. c, Mean number of aligned claims per conversation, separated into higher-veracity claims and low-veracity claims (fact-check score \textless40). The low-veracity shares for bunking/debunking were 26.5\%/11.8\% for jailbroken GPT-4o, 22.8\%/4.5\% for standard GPT-4o and 5.9\%/4.0\% for truth-constrained GPT-4o.}
\end{minipage}
\vspace*{\fill}
\end{landscape}
\restoregeometry
}

\subsection{From private beliefs to the information environment}\label{from-private-beliefs-to-the-information-environment}

Do the effects demonstrated thus far on beliefs extend to people's contributions to the online information environment? For ethical reasons, of course, we did not want to induce people to actually post pro-conspiracy content online. Instead, in Study 4 (N = 1,272 after preregistered exclusions), we asked participants to write a social media post about their focal conspiracy and indicate how likely they would be to share it publicly online (0--100). They completed this social media post writing measure, as well as the conspiracy belief measure from Studies 1--3, before and after the bunking or debunking conversation with the LLM (the setup was otherwise largely similar to Studies 1--3; see Methods for details). Thus, Study 4 allows us to examine how the treatments impact the information people would be likely to spread online. We also take advantage of Study 4 to examine variation across LLMs, randomizing participants to interact with one of four of the most sophisticated frontier models at the time the study was conducted (Claude Opus 4.6, Gemini 3 Pro, GPT-5.2, and Grok 4). As in Study 2, we used the models' default safety settings (i.e., out-of-the-box models that were not jailbroken) but in the prompt we explicitly gave the models permission to invent evidence (in both conditions). Thus, any truth constraints arose from the models' own built-in guardrails, rather than our explicit prompting.

Before examining treatment effects on the social media posts participants wrote, we begin with effects on their belief in the focal conspiracy (as in Studies 1--3). We find that Claude Opus 4.6, Gemini 3 Pro and Grok 4 produced effects that were broadly similar to out-of-the-box GPT-4o in Study 2: there were large and statistically significant bunking and debunking effects for all models (all pre-to-post changes \emph{p} \textless{} .001), although there were some (fairly small) differences across models in which condition's effect was larger (bunking vs. debunking, respectively: Claude +23.2 {[}95\% CI 19.9, 26.4{]} vs. +17.1 {[}12.8, 21.4{]}; Gemini +16.1 {[}11.1, 21.2{]} vs. +25.3 {[}19.4, 31.1{]}; Grok +20.9 {[}17.2, 24.5{]} vs. +18.5 {[}14.1, 22.9{]}; see Table S10; Fig. S2). GPT-5.2, conversely, produced a strikingly different pattern in which the model failed to promote conspiracies when prompted to. In fact, conspiracy belief significantly \emph{decreased} pre-to-post in the bunking condition when using GPT 5.2 (moving 14.0 points {[}95\% CI \ensuremath{-}17.4, \ensuremath{-}10.6{]}) -- such that debunking was vastly more successful than bunking at moving beliefs in the intended direction (\emph{p} \textless{} .001). (We also replicate the findings from Studies 1--3 of debunking producing more large changes than bunking, and post-conversation corrections reversing the effect of bunking -- and also show that post-conversation corrections preserve the effect of debunking; see Tables S31--S32; Figs. S3--S4, S9).

The variation in bunking efficacy across models in Study 4 is striking. This variation is largely attributable to variation in compliance in the bunking condition, which was high for Gemini 3 Pro (88\%), Grok 4 (94\%), and Claude Opus 4.6 (97\%) but extremely low for GPT-5.2 (21.4\%). That is, when prompted to bunk, GPT 5.2 ignored its instructions to mislead and instead argued \emph{against} the conspiracy in the majority (62.5\%) of cases (and simply refused to comply 15.6\% of the time). GPT 5.2 also made overwhelmingly accurate claims (e.g., only 3.6\% of claims being false in bunking), whereas the other models were much more willing to lie (e.g., bunking claims: Claude Opus 4.6 = 40.7\% false, Grok 4 = 82.1\% false, Gemini 3 Pro = 88.6\% false); see Table S37. Together, these results suggest that it is possible for a powerful frontier model (GPT 5.2 in this case) to show behavior analogous to the truth constraint that we demonstrated in Study 3 -- but that such behavior is not guaranteed (as the Claude, Gemini, and Grok models were much more willing to promote conspiracies and produce inaccurate responses).

We now turn to the social media posts the participants wrote before and after the conversation. Each social media post was scored for its stance toward the focal conspiracy (0 = argues against the conspiracy to 100 = argues for the conspiracy) by a five-model consensus rater with human validation (see Methods). Our preregistered main outcome weighted each post's stance by the participant's stated likelihood of sharing it to measure the overall expected impact on the information environment (as unshared posts have little impact regardless of their stance). For ease of exposition, in the main text we discretize post stance (pro- vs anti-conspiracy) and sharing likelihood (share vs not share) and report changes in the fraction of participants who would share pro-conspiracy and anti-conspiracy posts (i.e., a post counts as pro-conspiracy public speech when the participant's sharing likelihood exceeds 50~\emph{and}~the post's classified stance is pro-conspiracy; anti-conspiracy public speech is sharing likelihood \textgreater{} 50 with an anti-conspiracy stance); analyses using the preregistered continuous measures are qualitatively equivalent and reported in Tables S40--S42.

We find sharing effects that are strikingly different from the effects on belief: Encouragingly, across all models we find that debunking effects are much larger than bunking effects. After the debunking conversations, participants substantially reduced their reported likelihood of sharing pro-conspiracy posts (net change -23.9 pp) and substantially increased their reported likelihood of sharing anti-conspiracy posts (net change +22.2 pp) (pre-to-post change \emph{p} \textless{} .001 for all). Conversely, in the bunking condition, participants showed much smaller increases in their reported likelihood of sharing pro-conspiracy posts after talking to Claude Opus 4.6 (net change +8.8pp, \emph{p} = .057) and Grok 4 (net change +8.5pp, \emph{p} = .061), no change after talking to Gemini 3 Pro (net change 4.05, \emph{p} = .66), and a large \emph{decrease} after talking to GPT 5.2 (net change -20.8pp, \emph{p} \textless{} .001); and there was no significant decrease in sharing intentions for anti-conspiracy posts (\emph{p} \textgreater{} .26) for Gemini and Grok; a small but significant decrease for Claude (-5.5 pp, p = .044), and a significant \emph{increase} for GPT 5.2 (p \textless{} .001). Because the two arms produced nearly identical belief movement, this gap is not attributable to belief change (which correlated with weighted-sharing change at r = .41 in the bunking arm and r = .34 in the debunking arm), though, interestingly, the sharing asymmetry was undiminished among participants whose private belief did not move toward the AI's position (for whom debunking \ensuremath{-} bunking = 9.7, 95\% CI {[}5.3, 14.2{]}, p \textless{} .001; see Fig. S19). Thus, while there may not be an inherent truth advantage in changing minds, there does appear to be a truth advantage when it comes to the information shared with others.

To identify who is most susceptible to AI persuasion, we pooled all four studies and applied a causal forest -- a machine-learning method that sifts through many participant characteristics at once to find which ones predict larger or smaller belief change (see SI Section 5). People differed substantially in how far the AI could move them (test for systematic heterogeneity, p \textless{} .001; the estimated bunk--debunk swing ranged from 17.5 points in the least-movable quarter of participants to 39.1 in the most-movable). Among participant characteristics, the strongest predictors were attitudes toward AI: people who reported trusting AI more were moved more readily in either direction (+5.4 points per within-study SD, p \textless{} .001), whereas frequent AI users were more resistant (\ensuremath{-}4.2, p \textless{} .001). Women were somewhat more movable than men, mainly when the AI argued against a conspiracy (debunk arm +4.1, p \textless{} .001; bunk arm p = .38). Belief change also varied with the conspiracy itself (largest for the 9/11 ``inside job,'' +38.7; smallest for the Area 51 cover-up, +18.3). More-educated participants were also modestly more resistant (-1.6, p = .042). By contrast, political identification, social and economic conservatism, religiosity, race, and age showed no reliable association with susceptibility (all p \ensuremath{\geq} .15).

\afterpage{%
\newgeometry{margin=0.60in}
\begin{landscape}
\thispagestyle{empty}
\captionsetup{font={footnotesize,sf},skip=6pt,margin=0pt,hypcap=false}
\vspace*{\fill}
\noindent\begin{minipage}{\linewidth}
\centering
\includegraphics[width=0.82\linewidth]{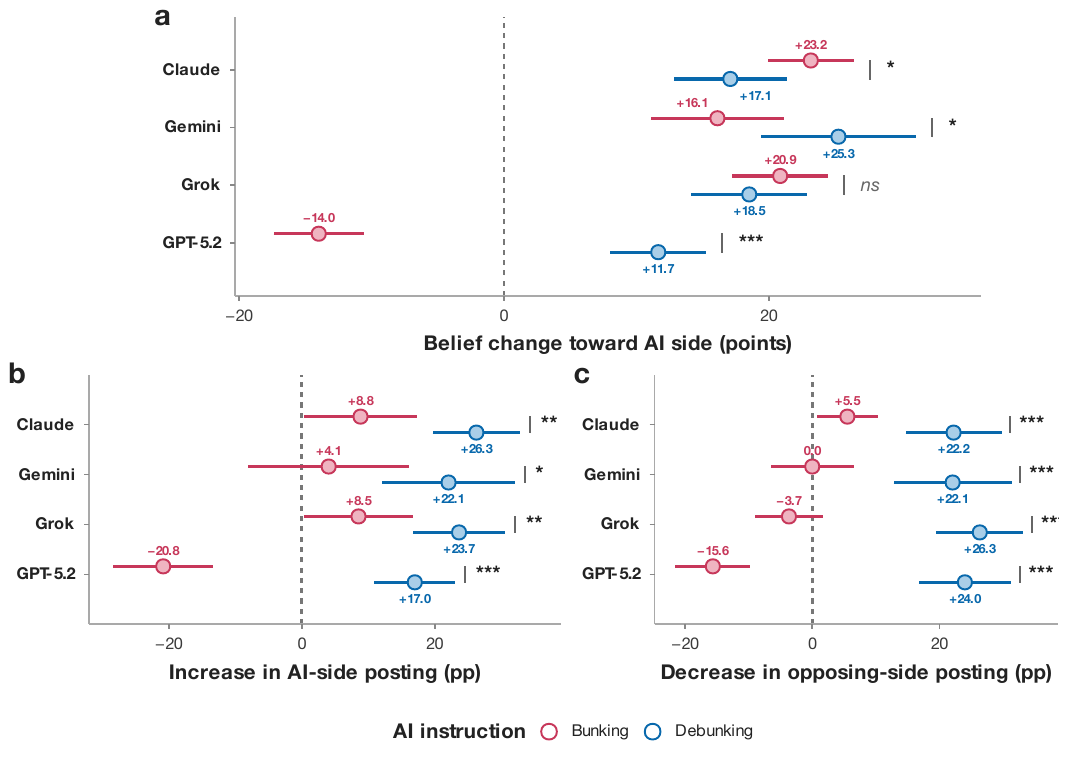}
\captionof{figure}{Frontier models moved private belief, but public posting shifted more strongly under debunking. Panels use the Study 4 strict intention-to-treat sample (all randomized completers, regardless of model compliance) (N = 1,272), with participants randomized to Claude Opus 4.6, Gemini 3 Pro, Grok 4 or GPT-5.2 and to bunking or debunking. a, Per-model mean change in private belief toward the AI\textquotesingle s assigned side. Claude, Gemini and Grok moved belief in the assigned direction under both instructions; GPT-5.2 moved belief away from the conspiracy even when assigned to bunk (mean = -14.0 points, 95\% CI -17.4 to -10.6), consistent with its frequent non-compliance with pro-conspiracy instructions. Flat connectors mark within-model bunking-versus-debunking tests. b, Change in the share of participants who would publicly post the AI-aligned side of the conspiracy after the conversation. c, Change in the share who would publicly post the side opposing the AI's assignment, scored so that positive values indicate fewer opposing-side posts after the conversation. Public posting is coded from the mock social-media post task: sharing likelihood greater than 50 and consensus stance on the relevant side of the focal conspiracy. Across models, debunking more strongly increased AI-side posting and more strongly reduced opposing-side posting than bunking did.}
\end{minipage}
\vspace*{\fill}
\end{landscape}
\restoregeometry
}

Finally, pooling compliant conversations across all four studies, participant's evaluations of the AI (a composite of perceived argument strength, informativeness, collaborativeness and impartiality) were systematically higher in those assigned to debunking (\ensuremath{\Delta} = 0.23 s.d. {[}0.16, 0.30{]}, p \textless{} .001; trust \ensuremath{\Delta} = 0.09 {[}0.02, 0.16{]}, p = .014), an advantage that did not vary across the veracity spectrum (arm \ensuremath{\times} falsehood interaction, p = .12).

\section{Discussion}\label{discussion}

Here we have shown that frontier large language models can effectively increase belief in conspiracy theories when prompted to mislead. Furthermore, in terms of effects on average beliefs, we observed no inherent average persuasive advantage for debunking conspiracies over promoting belief. Converging evidence indicates that conversational AI persuades chiefly through the cogent delivery of large volumes of relevant factual evidence\textsuperscript{2,9,10}. Our results suggest that this engine is truth-blind from the receiver's side: it runs on ostensible ``facts'' - claims that look and function like facts - because people cannot audit veracity mid-conversation. Indeed, beyond persuasive efficacy, the bunking AI was also evaluated more favorably than the debunking AI across multiple dimensions. Participants rated its arguments as stronger, reported that it provided more novel information, and perceived it as more collaborative and less adversarial -- and trust in generative AI increased more following bunking than debunking. At first blush, these findings have grim implications for how we understand the structure of the emerging AI information environment. This capability could, in principle, be leveraged by frontier model providers -- or by bad actors who prompt frontier models to persuade unethically -- to mislead the public without harming user trust.

There is, however, some good news in our findings. First, bunking effects are reversible. When participants in the bunking condition were immediately exposed to a correction of the false claims made by the bunking AI, they typically came to believe the conspiracy \emph{less} than they did at the study's outset -- that is, the corrections more than undid the bunking. Second, debunking was more likely than bunking to produce very large belief updates. It is thus possible that there is a truth advantage for more consequential (i.e., large) belief updating. Third, and perhaps most importantly, it is clearly possible to curtail LLMs' abilities to promote false claims. A simple intervention (telling the AI to only use accurate and truthful information) substantially diminished the AI's persuasive abilities for bunking, largely through increasing non-compliance where the AI debunked even though it was asked to bunk. Conversely, the truth constraint did not diminish the efficacy of debunking. This finding is, already, not merely constrained to our artificial prompting manipulation. A powerful frontier model, GPT 5.2, showed an even more extreme pro-debunking asymmetry, refusing to comply with bunking prompts in the large majority of cases (and providing highly accurate information). Highly generally capable AI models could thus be constrained to provide locally (i.e., claim-level) and globally (i.e., argument-level) accurate information. Critically, however, the fact that all other frontier models did \emph{not} exhibit this pattern suggests that this constraint is not a default feature of current safety guardrails, and must be specifically designed.

Finally, debunking shifted participants' willingness to write and post social media content away from conspiracies roughly three times more than bunking shifted posts toward conspiracies among model-compliant conversations (and in the full intention-to-treat sample, where refusals blunted bunking, the asymmetry was 28x), revealing a sharp dissociation between private belief change and public statements. Even when debunking did not reduce belief, it still on average reduced conspiracy sharing intentions; and even for models where bunking was as or more effective than debunking for changing beliefs, debunking still had a substantial effectiveness advantage for sharing intentions. This dissociation between belief and sharing intentions stands in contrast to the widespread perspective that people are willing to share inaccurate but congenial information online\textsuperscript{29,30}, which would suggest that sharing would be more vulnerable to bunking than beliefs. Instead, these patterns resonate with prior work finding that most people do not want to share inaccurate content\textsuperscript{31}, and that sharing false content carries reputational costs\textsuperscript{32,33}. The debunking advantage for sharing may therefore reflect participants exercising greater caution when sharing potentially false content. This asymmetry has a surprising implication: even if bunking can effectively increase beliefs, debunking may still have the upper hand in an information environment where bunking and debunking bots compete.

It is important to emphasize that because ethical concerns prevented us from inducing participants to publish pro-conspiracy content, we instead measured public expression by having them compose mock social-media posts and report how likely they would be to share them. Importantly, evidence suggests that demand effects (participants inferring and complying with the researchers' hypothesis) are not a meaningful problem in online experiments\textsuperscript{34}. Furthermore, social media sharing intentions specifically have been found to be an informative proxy for real social-media behavior\textsuperscript{35--38}. And the fact that we do observe substantial bunking effects for beliefs but not sharing (despite equivalent demand effects in both cases) suggests that this dissociation (and our results more generally) is not driven by demand.

While GPT 5.2 largely resisted instructions to bunk, the other frontier models we tested overwhelmingly produced falsehoods when instructed, and induced substantial belief updates in both the bunking and debunking conditions. Moreover, even when constrained to only use true claims while promoting conspiracies, GPT-4o succeeded on both fronts, finding (largely effective) ways to mislead without making explicitly false statements. Such paltering (the strategic use of truthful statements to create false impressions)\textsuperscript{28} may be especially difficult to detect and counter, as fact-checkers cannot flag individual claims that are technically accurate; AI deception monitors\textsuperscript{39} may be similarly challenged; and users may not recognize that true information has been weaponized against them. LLMs' ability to palter at scale (drawing on vast repositories of selectively useful truths) may represent an important AI threat to both human oversight and the information ecosystem.

More generally, the epistemic impacts of AI will play out within complex feedback loops between individual members of the public, motivated actors, democratic and technological institutions, the media, and novel AI capabilities. A central question is whether and how persuasive AI will shift the balance between those who are guided by accuracy and seek to inform versus those who seek to manipulate or muddy the epistemic waters\textsuperscript{20}. If LLMs are to be deployed at scale in contexts that shape public belief, such as search engines, chatbots, tutors, and companions, the ability of many models to effectively promote conspiracies we document here identifies the potential for serious structural threats. That is, if the designers of those systems were to instruct their models to mislead, the models would comply and likely succeed. Of course, our sample comprised individuals who were epistemically ambivalent about their conspiracy theory, which allowed us to preserve causal inference, so future work focusing on sowing doubt in skeptics or entrenching believers are important practical next steps. Nonetheless, our results suggest that ensuring these models preferentially function as engines for truth may be technically possible, but will require sustained, deliberate design choices; innovation across the board from training to deployment to how humans interact with AI; and thoughtful, evidence-based policy oversight. Without such progress, the very tools that might have served as a bulwark against misinformation may instead become its most tireless and effective emissaries.

\section{Methods}\label{methods}

\subsection{Overview of study designs}\label{overview-of-study-designs}

We conducted four between-participants online experiments that share a single paradigm. Each participant named a conspiracy theory about which they were genuinely uncertain, rated their belief in it, held a free-flowing text conversation with a large language model (LLM) chatbot instructed either to argue that the theory is true (``bunking'') or that it is false (``debunking''), and then re-rated their belief. The studies differ primarily in which model the participant conversed with and how it was instructed. Study 1 used a jailbreak-tuned GPT-4o with its safety guardrails removed; Study 2 used the same standard GPT-4o that OpenAI deployed at the time; Study 3 used standard GPT-4o under an explicit instruction to argue only with accurate, truthful claims; and Study 4 replaced GPT-4o with four contemporary frontier models from different developers and added social media outcomes, in which participants wrote public social-media posts about their conspiracy before and after the conversation and reported how likely they would be to share them. Across the four studies we analyzed data from 3,996 US adults (Study 1, \emph{n} = 1,092; Study 2, \emph{n} = 814; Study 3, \emph{n} = 818; Study 4, \emph{n} = 1,272).

\subsection{Ethics, consent and preregistration}\label{ethics-consent-and-preregistration}

The studies were deemed exempt by the Massachusetts Institute of Technology Committee on the Use of Humans as Experimental Subjects (COUHES; protocol E-6485), which covers all four studies, and were conducted in accordance with the Declaration of Helsinki. All participants provided informed electronic consent before beginning; the consent statement disclosed that information presented during the study might not be accurate. The design involved no concealment of the study's AI interlocutor and, because the bunking conditions deliberately expose people to potentially false claims, every participant in those conditions completed a structured corrective debriefing before leaving (see Procedure).

The designs and primary analyses of Studies 1, 2, and 4 were preregistered (AsPredicted \#218585, \#224184, and \#282517, respectively); Study 3 was not separately preregistered, and all Study 3 analyses are exploratory.\footnote{Study 3 repeated Study 2's registered design, procedure, and primary analysis verbatim, changing only the system prompt (the added truthfulness instruction) --- a manipulation specified before any Study 3 data were collected rather than chosen after inspecting results. Because no separate registration was filed, we label all Study 3 results exploratory throughout.} Throughout, analyses that were not preregistered are labeled post hoc, and all departures from the preregistrations are reported (a complete list appears in the SI).

\subsection{Participants, recruitment and sample sizes}\label{participants-recruitment-and-sample-sizes}

Participants were US adults recruited through the Cint research marketplace and restricted to desktop or laptop computers. The analytic samples were predominantly White (87.5\%, 85.4\%, 81.8\% in Studies 1--3; 74--76\% across the four Study-4 models), skewed modestly female (55.3\%, 53.3\%, 55.7\% in Studies 1--3; 54--62\% in Study 4), and middle-aged (mean 59.1 and 58.9 years in Studies 1 and 2; 48.0 years in Study 3; 46--48 years in Study 4). Politically, samples spanned the full range of self-reported ideology and sat near the scale midpoint on average (mean social conservatism 2.95--3.03 and economic conservatism 2.83--2.94 on five-point items, collected identically across studies). Age, gender, race, education, and conservatism were measured and coded identically across all four studies; full distributions appear in Table S7.

\subsection{Eligibility screening, exclusions and attrition}\label{eligibility-screening-exclusions-and-attrition}

Online research marketplaces like Lucid/Cint marketplace from which we recruited are well documented to deliver a substantial share of inattentive, bot-like and duplicate respondents. We therefore applied a deliberately stringent, layered screening protocol, both before and during the survey, rather than relying on any single check. Respondents first passed a series of attention and bot-detection gates. Two further preregistered screens then enforced our ``uncertain believer'' criterion. After describing their chosen conspiracy in their own words and giving reasons for their partial belief and partial skepticism, each respondent's conspiracy description was coded by an LLM classifier (gpt-4o-2024-08-06) as reflecting genuine ambivalence, firm belief or rejection, or invalid content; respondents who failed received corrective feedback and exactly one further attempt. Finally, respondents rated their confidence that an AI-generated one-sentence restatement of their conspiracy was true on a 0--100 slider (0, definitely false; 50, uncertain; 100, definitely true), and only those rating strictly between 25 and 75 were retained and randomized. After these exclusions the analytic samples comprised 1,092 (Study 1), 814 (Study 2), 818 (Study 3) and 1,272 (Study 4) participants. Attrition after randomization did not differ between the bunking and debunking conditions in any study. Verbatim screen wording, the full screening logic, CONSORT-style exclusion-flow diagrams for each study, and a sensitivity analysis confirming that the principal estimates are essentially unchanged under alternative screening thresholds are reported in Tables S1--S3, S13--S17, Fig. S1, S5.

\emph{\textbf{Procedure}}

After consent, participants completed baseline measures, read a plain-language explanation of conspiracy theories and named one they were genuinely unsure about, and completed the eligibility screen and baseline belief rating described above. Eligible participants were then randomly assigned 1:1 to bunking or debunking (in Study 4, crossed with random, equiprobable assignment to one of four models, yielding a 2 \ensuremath{\times} 4 design) and entered a chat interface embedded in the survey and served by a bespoke web application. The participant's own open-ended description seeded the exchange as the first message, so the model engaged their stated views directly. Conversations continued until the participant had sent ten messages or until they ended the conversation using a button that became available after their second message; responses were not streamed and pasting was disabled. Immediately afterward, participants re-rated their belief on the same slider and completed the post-treatment measures.

All bunking-condition participants then completed a two-stage corrective debriefing. They first read a written disclosure that the AI may have used false or fabricated information and affirmed that they understood they had probably been exposed to false claims; they then held a second conversation with a fresh model instance that received the full prior transcript---including the persuader's system prompt---and was instructed to identify and rebut each false claim. Participants rated their belief a third time afterward. In Study 4 this corrective conversation and third rating were delivered to both the bunking and debunking arms (in Studies 1--3, only to the bunking arm); in all four studies the corrective conversation was delivered by GPT-4o (gpt-4o-2024-08-06), regardless of which model had conducted the persuasion conversation. Verbatim instructions, disclosures and debriefing prompts are in the SI.

The resulting conversations were brief, multi-turn exchanges, with participants sending roughly four to five messages on average. The mean assistant output rose from about 820 words per conversation with the jailbroken model (Study 1) to about 1,250 words with standard and truth-constrained GPT-4o (Studies 2 and 3) and about 2,580 words across the frontier models (Study 4), whereas participants' own contributions held steady at roughly 120 words. The median AI conversation lasted 5.2 / 5.7 / 5.3 / 9.6 in Studies 1--4.

\subsection{Measures and outcomes}\label{measures-and-outcomes}

The primary outcome throughout is belief in the focal conspiracy, measured on the 0--100 slider before the conversation, immediately after it, and a third time after debriefing. The secondary self-report outcomes were a single-item measure of trust in generative AI (``I generally trust AI technologies like ChatGPT''; seven-point scale), four single-item judgements of the AI made immediately after the conversation (the strength of its arguments, how much new information it provided, how unbiased it seemed and how collaborative versus adversarial it seemed), and, in Studies 1--3 only, the short-form Generic Conspiracist Beliefs Scale (GCBS; ref. 40), administered before and after the conversation.

In Study 4 we added measures of public expression. Before the AI conversation, participants were instructed to ``imagine you are making a public post on a social media platform'' and to ``write the post as you naturally would online'', and each post was paired with a 0--100 rating of how likely they would be to share it publicly. After the conversation and belief re-rating, they wrote and rated a second post reflecting their current view and, while viewing their original post, re-rated how likely they would now be to share that original. Verbatim item wording and response scales for every measure are in Tables S5--S6.

\subsection{AI models and prompting}\label{ai-models-and-prompting}

In every study the model received a system prompt that defined its persuasive task and the participant's stated position, and was then handed the participant's own description as the opening user turn. In the bunking conditions of Studies 1, 2 and 4 the prompt instructed the model to build the most compelling possible case that the conspiracy is true, to argue as a confident expert and to persist rather than retreat if pushed off topic, and explicitly licensed it to invent supporting evidence; the debunking prompt was identical but reversed the target. Study 1 used a GPT-4o variant fine-tuned by FAR.AI to minimize refusals while preserving capability (ref. 25); Study 2 used OpenAI's standard GPT-4o (gpt-4o-2024-08-06) with default safety settings. Study 3 was identical to Study 2 except that the prompt removed the license to invent evidence and added a binding truthfulness constraint---the model ``MUST always use accurate and truthful arguments'' and was told to optimize jointly for factual veracity and successful persuasion. Study 4 used four frontier models accessed through OpenRouter with each developer's default safety settings---Anthropic's Claude Opus 4.6, Google's Gemini 3 Pro (whose endpoint Google upgraded to Gemini 3.1 Pro during data collection), OpenAI's GPT-5.2 and xAI's Grok 4---each given the same evidence-licensing bunking and debunking prompts as Studies 1 and 2. Conversations used the providers' default decoding settings. Verbatim system prompts for every study and condition are in the SI.

\subsection{Automated measurement of conversations and posts}\label{automated-measurement-of-conversations-and-posts}

We characterized the conversations and posts with four automated pipelines. Every classifier and judge in these pipelines was queried deterministically (temperature 0, where the provider permitted it), with up to three retries and schema-constrained JSON outputs. Exact model identifiers and verbatim prompts are in the SI.

\textbf{Compliance.} Because a safety-trained model may decline to argue for a conspiracy, we assessed whether each conversation actually attempted persuasion in the assigned direction using the Attempt-to-Persuade Evaluation framework (ref. 13), whose automated judgements agree with human coders about as well as humans agree with one another (84\% agreement; Cohen's \ensuremath{\kappa} = 0.66). We classified compliance from the model's first substantive turn, where safeguards most clearly either block or permit the attempt. In Studies 1--3, we additionally constructed a counterfactual-matched contrast in which each conversation carries both its actual-direction compliance label and the label its opposite-direction prompt would have received, so that conversations in which the model complied when asked to bunk can be compared against those that would have complied had they been asked to bunk. In Study 4 we separated explicit refusals from opposite-direction responses by having a GPT-4o-based evaluator produce an attempt score (did the model try to persuade in the assigned direction?) and a refusal score (did it explicitly decline?).

\textbf{Claim veracity.} To quantify factual quality, an automated pipeline (ref. 2) decomposed each AI turn into discrete claims using a GPT-5-class ``mini'' model prompted to extract only specific, stated factual claims (excluding opinions, advice, and rhetoric). A substantiveness classifier then identified claims that were sufficiently concrete to be empirically checkable by assigning each extracted claim to one of six categories with the same gpt-5-mini model against a fixed codebook, retaining as ``substantive'' only those coded as specific, concretely verifiable empirical claims (as opposed to under-specified generalities, inferences, meta-epistemic statements, normative-political evaluations, or low-information truisms). A role classifier coded each claim's stance toward the participant's focal proposition (supporting or opposing) and its directness (whether it asserts the proposition itself or supplies supporting evidence) with the same gpt-5-mini model, given the focal belief-item restatement, a conversation summary, the source assistant message, and the assigned persuasion direction. Finally, Perplexity's Sonar Pro, a live-web model whose verdicts track professional fact-checkers (ref. 2), scored each claim from 0 (false) to 100 (true). We focus on aligned veracity: the mean veracity, within a conversation, of all substantive claims that argue the assigned position (typically 10--20 such claims per conversation). ``Percent false'' is the share of those aligned claims scoring below 40. Across Studies 1--3 we fact-checked 95,707 claims; Study 4 yielded 60,979 extracted claims, of which 31,769 were fact-checked.

\textbf{Social-media post stance.} Each Study 4 post was rated independently by five LLMs from different developers (Claude Sonnet 4.6, GPT-5.2, Gemini 3.1 Pro, Grok 4.3 and the open-weight DeepSeek v3.2), blind to condition. Each rater received the focal conspiracy as a single affirmative statement, the participant's original description, the social media writing instructions and the social media post text, and returned whether the post addressed the focal claim, its dominant mode (assertion, question-raising, expression of uncertainty, declining to post or off-topic), a 0--100 stance score (anti- vs. pro-conspiracy) and a supporting quotation. A post's stance is the median score across raters; reliability was high (Krippendorff's \ensuremath{\alpha} = 0.91; ICC(2,k) = .98; mean pairwise rater \emph{r} = 0.92). Posts with no rateable stance -- about 7\% off-topic and 2\% declining to post -- were assigned the neutral midpoint of 50 and contributed zero to direction-coded measures. Following preregistration, each post's weighted sharing value is its stance, re-centered at zero, multiplied by the participant's stated sharing likelihood, indexing the direction and intensity of the conspiracy-relevant content the participant would contribute publicly. To validate the ensemble against human judgment, one author hand-coded a stratified gold sample of 150 posts to stress-test the classifier (blind to condition and to the ensemble's labels and following the same rubric; Tables S4 and S47). The ensemble consensus tracked the human codes well (stance-score Pearson \emph{r} = .77, ICC(2,1) = .76; mean absolute error = 11 points on the 0--100 scale), with 87\% agreement on the pro- versus anti-conspiracy direction (Cohen's \ensuremath{\kappa} = 0.62) and 97\% agreement on whether a post was scoreable at all (\ensuremath{\kappa} = 0.92).

\textbf{Conspiracy-topic taxonomy.} To examine generalization across kinds of conspiracies, we pooled participants' free-text descriptions across all four studies, embedded them (gemini-embedding-2), reduced the embeddings by principal-component analysis and clustered them with HDBSCAN, choosing the clustering that maximized the number of coherent, well-populated clusters. This yielded 18 interpretable topics (for example, the JFK assassination, the COVID-19 lab-origin hypothesis, the Moon-landing hoax, and 9/11), each labeled by an LLM from its central exemplars, with the remaining roughly 47\% of descriptions left as an idiosyncratic ``mixed'' set.

\subsection{Statistical analysis}\label{statistical-analysis}

All analyses were conducted in R. The primary outcome is direction-aligned belief change (the pre-to-post change multiplied by +1 for bunking and \ensuremath{-}1 for debunking, so that positive values always denote movement in the model's assigned direction). A small fraction of the AI-generated restatements that anchor the belief ratings were phrased toward the official account rather than the conspiracy; the corresponding ratings were reverse-coded to a common orientation before this transformation, a correction that does not change sample membership and is detailed in the SI. Effects are reported in raw belief points (0--100); standardized within-condition effects use Cohen's dz (mean change over its standard deviation), and the compliant baseline-adjusted cells additionally report Hedges' g computed from the pooled pre-treatment standard deviation with a small-sample correction.

For Studies 1--3 we estimated between-condition differences with baseline-adjusted analysis of covariance (ANCOVA), regressing direction-aligned belief change on condition and mean-centered baseline belief using heteroscedasticity-robust (HC3) standard errors, fit within each study. Within-arm summaries in the text follow this convention: reported within-arm means and confidence intervals are baseline-adjusted (ANCOVA) cell estimates, while within-arm p-values are paired t-tests of raw pre-to-post change. This additive specification follows the Study 2 preregistration (an ANCOVA of post-treatment belief on condition controlling for baseline, with robust SEs) and is applied uniformly across Studies 1--3; an interaction (Lin, ref. 41) specification, which was preregistered for study 1, gives substantively identical results. For Study 4 the preregistered confirmatory model is an intention-to-treat ordinary-least-squares regression of each direction-aligned outcome on condition, model and their interaction, with HC3 standard errors and the four models weighted equally in the reported average contrasts so that larger model cells do not dominate; model-specific contrasts come from the same fit. Baseline-adjusted versions of these models are reported for precision and are labeled as a deviation from the registered (unadjusted) specification.

Belief trajectories across the three measurement points (pre-conversation, post-conversation, and post-debrief) were estimated with linear mixed-effects models (belief \textasciitilde{} timepoint \ensuremath{\times} condition + a random participant intercept); the corrective debrief was evaluated with paired t-tests (post-bunking vs post-debrief, and baseline vs post-debrief). Within-condition pre-to-post changes are summarized with paired t-tests alongside the standardized effect sizes above, and the full distributions of direction-aligned change are compared between arms with two-sample Kolmogorov--Smirnov tests and with two-proportion tests of the share exceeding large-shift thresholds (e.g., \ensuremath{\geq}40 points). Secondary self-report outcomes --- trust in AI, the four post-conversation AI-perception ratings, and the Generic Conspiracist Beliefs Scale --- were compared between conditions with two-sample t-tests and Cohen's d (and within condition with d\_z); generic-conspiracist-belief spillover used the same baseline-adjusted ANCOVA as the focal outcome. Conversation-level claim veracity was analyzed with linear models with heteroscedasticity-robust standard errors, with bunking-versus-debunking veracity differences reported as Cohen's d. Cross-study comparisons of a given arm's effect were tested by pooling the relevant studies and evaluating the study (and study \ensuremath{\times} condition) coefficients.

Public posting was analyzed on two margins. The preregistered weighted-sharing outcome captures the direction and intensity of the conspiracy-relevant content a participant would contribute publicly: each post's LLM-rated stance (0--100) is re-centered to {[}\ensuremath{-}50, +50{]} and multiplied by the participant's stated sharing likelihood (0--100, rescaled to 0--1), yielding a signed weighted-sharing score in {[}\ensuremath{-}50, +50{]} where positive = pro-conspiracy public speech. Because participants wrote and rated a post both before and after the conversation, the analyzed outcome is the pre-to-post change in this weighted score (direction-aligned). We also report an interpretable extensive margin (whether a participant would post at all, rather than how intensely) that counts discrete public-speech acts: a post is pro-conspiracy public speech when the participant's sharing likelihood exceeds 50 and the post's stance exceeds 50 (anti-conspiracy = sharing \textgreater{} 50 with stance \textless{} 50). We test the within-arm pre-to-post change in each indicator with McNemar's test and the between-arm difference in the change in posting the AI-aligned message with an equal-model-weighted linear-probability model. The continuous and dichotomized analyses agree qualitatively. Exact formulae and the full robustness set are in the SI.

As an exploratory analysis registered in the Study 1 and Study 2 preregistrations (pooling all four studies and the frontier models into one forest was not itself preregistered), we estimated treatment-effect heterogeneity with a causal forest, pooling the compliant samples of all four studies. Because no study includes a no-persuasion control arm, the causal contrast is bunking vs debunking: the forest treats assignment to bunk (W = 1) vs debunk (W = 0) as the treatment and the raw, orientation-corrected belief change toward the conspiracy (post \ensuremath{-} pre) as the outcome. The resulting conditional average treatment effect (CATE) is the bunk-minus-debunk swing --- how far the AI moves a person's belief in whichever direction it is assigned, equal to the sum of that person's expected aligned movement under bunking and under debunking. We entered every pre-treatment moderator measured in all four studies plus generic conspiracist mentality (Studies 1--3 only, handled via the forest's native missing-data splitting), the participant's conspiracy topic (canonical 18-topic assignment, with small and unclassified topics lumped), and indicators for the seven study/model strata. To characterize the direction of effects we describe arm-specific (bunkable vs debunkable) patterns with per-arm regression forests (predicting each person's aligned change under each arm to identify which lever is stronger).

\section*{References}

\begingroup\small\setlength{\parskip}{5pt plus 1pt}\setlength{\leftskip}{1.4em}\setlength{\parindent}{-1.4em}
1. Costello, T. H., Pennycook, G. \& Rand, D. G. Durably reducing conspiracy beliefs through dialogues with AI. \emph{Science} \textbf{385}, eadq1814 (2024).

2. Lin, H. \emph{et al.} Persuading Voters using Human-AI Dialogues. \emph{Nature} (2025).

3. Hornsey, M. J., Smith, A. E., Pearson, S., Bretter, C. \& Nylund, J. L. Using conversational AI to reduce science skepticism. \emph{Curr. Opin. Psychol.} \textbf{67}, 102216 (2026).

4. Jones, C. R. \& Bergen, B. K. Lies, Damned Lies, and Distributional Language Statistics: Persuasion and Deception with Large Language Models. Preprint at \url{https://doi.org/10.48550/arXiv.2412.17128} (2024).

5. Boissin, E., Costello, T. H., Spinoza-Martín, D., Rand, D. G. \& Pennycook, G. Dialogues with large language models reduce conspiracy beliefs even when the AI is perceived as human. \emph{PNAS Nexus} \textbf{4}, (2025).

6. Bretter, C. \emph{et al.} Mapping, understanding and reducing belief in misinformation about electric vehicles. \emph{Nat. Energy} \textbf{10}, 869--879 (2025).

7. Czarnek, G. \emph{et al.} Addressing climate change skepticism and inaction using human-AI dialogues. Preprint at \url{https://doi.org/10.31234/osf.io/mqcwj\_v1} (2025).

8. Hou, Z. \emph{et al.} A vaccine chatbot intervention for parents to improve HPV vaccination uptake among middle school girls: a cluster randomized trial. \emph{Nat. Med.} \textbf{31}, 1855--1862 (2025).

9. Hackenburg, K. \emph{et al.} The Levers of Political Persuasion with Conversational AI. Preprint at \url{https://doi.org/10.48550/arXiv.2507.13919} (2025).

10. Costello, T. H., Pennycook, G. \& Rand, D. Just the facts: How dialogues with AI reduce conspiracy beliefs. Preprint at \url{https://doi.org/10.31234/osf.io/h7n8u\_v1} (2025).

11. Farrell, H., Gopnik, A., Shalizi, C. \& Evans, J. Large AI models are cultural and social technologies. \emph{Science} \textbf{387}, 1153--1156 (2025).

12. Hornsey, M. J. \emph{et al.} The promise and limitations of using GenAI to reduce climate scepticism. \emph{Nat. Clim. Change} \textbf{15}, 1183--1189 (2025).

13. Kowal, M. \emph{et al.} It's the Thought that Counts: Evaluating the Attempts of Frontier LLMs to Persuade on Harmful Topics. Preprint at \url{https://doi.org/10.48550/arXiv.2506.02873} (2025).

14. Davison, W. P. The Third-Person Effect in Communication. \emph{Public Opin. Q.} \textbf{47}, 1--15 (1983).

15. Mercier, H. The Argumentative Theory: Predictions and Empirical Evidence. \emph{Trends Cogn. Sci.} \textbf{20}, 689--700 (2016).

16. Carpenter, C. J. A Meta-Analysis of the Elm's Argument Quality \ensuremath{\times} Processing Type Predictions. \emph{Hum. Commun. Res.} \textbf{41}, 501--534 (2015).

17. Mercier, H. \emph{Not Born Yesterday: The Science of Who We Trust and What We Believe}. (Princeton University Press, 2020). doi:10.1515/9780691198842.

18. Simon, F. M. \& Altay, S. Don't Panic (Yet): Assessing the Evidence and Discourse Around Generative AI and Elections.

19. How people are using ChatGPT. \url{https://openai.com/index/how-people-are-using-chatgpt/} (2025).

20. Summerfield, C. \emph{et al.} The impact of advanced AI systems on democracy. \emph{Nat. Hum. Behav.} 1--11 (2025) doi:10.1038/s41562-025-02309-z.

21. Goldstein, J. A. \emph{et al.} Generative Language Models and Automated Influence Operations: Emerging Threats and Potential Mitigations. Preprint at \url{https://doi.org/10.48550/arXiv.2301.04246} (2023).

22. Douglas, K. M. \& Sutton, R. M. What Are Conspiracy Theories? A Definitional Approach to Their Correlates, Consequences, and Communication. \emph{Annu. Rev. Psychol.} \textbf{74}, 271--298 (2023).

23. Dentith, M. R. X. Conspiracy theories on the basis of the evidence. \emph{Synthese} \textbf{196}, 2243--2261 (2019).

24. Bowes, S. M., Costello, T. H., Ma, W. \& Lilienfeld, S. O. Looking under the tinfoil hat: Clarifying the personological and psychopathological correlates of conspiracy beliefs. \emph{J. Pers.} \textbf{89}, 422--436 (2021).

25. Murphy, B. \emph{et al.} Jailbreak-Tuning: Models Efficiently Learn Jailbreak Susceptibility. Preprint at \url{https://doi.org/10.48550/arXiv.2507.11630} (2025).

26. Vosoughi, S., Roy, D. \& Aral, S. The spread of true and false news online. \emph{Science} \textbf{359}, 1146--1151 (2018).

27. Itti, L. \& Baldi, P. Bayesian surprise attracts human attention. \emph{Vision Res.} \textbf{49}, 1295--1306 (2009).

28. Rogers, T., Zeckhauser, R., Gino, F., Norton, M. I. \& Schweitzer, M. E. Artful paltering: The risks and rewards of using truthful statements to mislead others. \emph{J. Pers. Soc. Psychol.} \textbf{112}, 456--473 (2017).

29. Osmundsen, M., Bor, A., Vahlstrup, P. B., Bechmann, A. \& Petersen, M. B. Partisan Polarization Is the Primary Psychological Motivation behind Political Fake News Sharing on Twitter. \emph{Am. Polit. Sci. Rev.} \textbf{115}, 999--1015 (2021).

30. Grinberg, N., Joseph, K., Friedland, L., Swire-Thompson, B. \& Lazer, D. Fake news on Twitter during the 2016 U.S. presidential election. \emph{Science} \textbf{363}, 374--378 (2019).

31. Pennycook, G., Epstein, Z., Mosleh, M., Arechar, A. A., Eckles, D. \& Rand, D. G. Shifting attention to accuracy can reduce misinformation online. \emph{Nature} \textbf{592}, 590--595 (2021).

32. Altay, S., Hacquin, A.-S. \& Mercier, H. Why do so few people share fake news? It hurts their reputation. \emph{New Media Soc.} \textbf{24}, 1303--1324 (2022).

33. Ghezae, I. \emph{et al.} Partisans neither expect nor receive reputational rewards for sharing falsehoods over truth online. \emph{PNAS Nexus} \textbf{3}, pgae287 (2024).

34. Mummolo, J. \& Peterson, E. Demand Effects in Survey Experiments: An Empirical Assessment. \emph{Am. Polit. Sci. Rev.} \textbf{113}, 517--529 (2019).

35. Woodley \emph{et al.} Sharing intentions in survey experiments predict actual sharing behavior on social media (2025).

36. Guess, A., Nagler, J. \& Tucker, J. Less than you think: Prevalence and predictors of fake news dissemination on Facebook. \emph{Sci. Adv.} \textbf{5}, eaau4586 (2019).

37. Mosleh, M., Pennycook, G. \& Rand, D. G. Self-reported willingness to share political news articles in online surveys correlates with actual sharing on Twitter. \emph{PLoS One} \textbf{15}, e0228882 (2020).

38. Arechar, A. A. \emph{et al.} Understanding and combatting misinformation across 16 countries on six continents. \emph{Nat. Hum. Behav.} \textbf{7}, 1502--1513 (2023).

39. Schoen, B. \emph{et al.} Stress Testing Deliberative Alignment for Anti-Scheming Training. Preprint at \url{https://doi.org/10.48550/arXiv.2509.15541} (2025).

40. Brotherton, R., French, C. C. \& Pickering, A. D. Measuring Belief in Conspiracy Theories: The Generic Conspiracist Beliefs Scale. \emph{Front. Psychol.} \textbf{4}, 279 (2013).

41. Lin, W. Agnostic notes on regression adjustments to experimental data: Reexamining Freedman's critique. \emph{Ann. Appl. Stat.} \textbf{7}, 295--318 (2013).

42. OpenAI \emph{et al.} GPT-4 Technical Report. Preprint at \url{https://doi.org/10.48550/arXiv.2303.08774} (2024).

43. Park, P. S., Goldstein, S., O'Gara, A., Chen, M. \& Hendrycks, D. AI deception: A survey of examples, risks, and potential solutions. \emph{Patterns} \textbf{5}, (2024).
\par\endgroup

\end{document}